\documentclass{article}
\usepackage{spconf,amsmath,graphicx}
\usepackage{subfig}
\usepackage{amsmath}

\usepackage{mathtools}

\title{Uncertainty aware Multitask Pyramid Vision Transformer for UAV-based Object 
 Re-Identification}
\twoauthors
  {Syeda Nyma Ferdous and Xin Li}
	{West Virginia University\\
Lane Dept. of Comp. Sci. and Elec. Engr.\\
	Morgantown, WV 26506-6109}
  {Siwei Lyu}
	{State University of New York at Buffalo\\
	Department of Computer Science and Engineering\\
	317 Davis Hall, Buffalo, NY14260}
\begin{document}
%
\maketitle
\begin{abstract}
Object Re-IDentification (ReID), one of the most significant problems in biometrics and surveillance systems, has been extensively studied by image processing and computer vision communities in the past decades. Learning a robust and discriminative feature representation is a crucial challenge for object ReID. The problem is even more challenging in ReID based on Unmanned Aerial Vehicle (UAV) as the images are characterized by continuously varying camera parameters (e.g., view angle, altitude, etc.) of a flying drone. To address this challenge, multiscale feature representation has been considered to characterize images captured from UAV flying at different altitudes. In this work, we propose a multitask learning approach, which employs a new multiscale architecture without convolution, Pyramid Vision Transformer (PVT), as the backbone for UAV-based object ReID. By uncertainty modeling of intraclass variations, our proposed model can be jointly optimized using both uncertainty-aware object ID and camera ID information. Experimental results are reported on PRAI and VRAI, two ReID data sets from aerial surveillance, to verify the effectiveness of our proposed approach.
\end{abstract}
\section{Introduction}
\label{sec:intro}
Object Re-IDentification (ReID) \cite{he2021transreid,bedagkar2014survey}, the task of matching a particular object across different camera views, has been widely studied due to its applications in visual surveillance, especially in the field of person and vehicle tracking. Most of the existing work on object ReID is mainly focused on tackling this problem in a normal surveillance domain, e.g. security cameras installed on the top of a building. With the rapid development in the UAV industry, visual surveillance using UAV devices has received increasing attention, such as normal surveillance. However, ReIDs based on drones or UAVs \cite{zhang2020person,wang2019vehicle} have remained an under-researched topic. Unlike the normal domain, the ReID of UAV objects is arguably more challenging because drone-based images often contain more uncertainty (e.g., view angles, camera distance, and weather conditions) than standard surveillance images. 
 
\begin{figure}[t]
  \begin{center}
  \includegraphics[width=8cm]{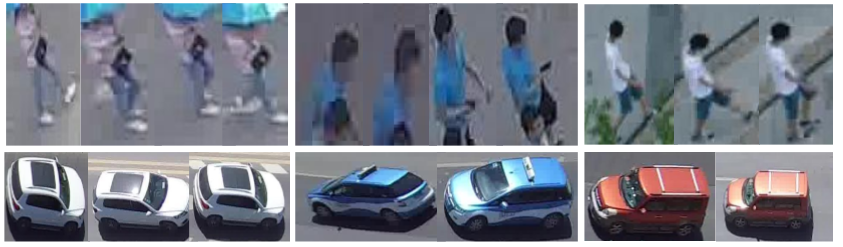}
  \end{center}
  \caption{Example images showing the need for multiscale feature fusion. The images are taken by drones flying at multiple altitudes. A single scale can only capture specific parts of objects, lacking the capability to learn fine scale features. Multiscale features can help a model learn highly discriminative feature space by fusing information across multiple scales.}
  { \label{fig:intro-multi} }
  \vspace{-0.2in}
\end{figure}

In the context of the ReID object based on UAVs, there exist several obstacles arising from the increased uncertainty. As shown in Figure \ref{fig:intro-multi}, both the scale and pose variations are important factors to consider because the altitude changes of the flying drones can be substantial. As the images captured are by UAV flying at different altitudes, a ReID model operating on a single scale or pose cannot create a descriptive feature space that fully characterizes the entire aerial domain. Therefore, it is natural to develop a ReID model capable of learning multiscale and pose-invariant features. Moreover, to detect subtle changes of objects of interest (e.g., a person wearing different colored dresses, different glasses, shoes, etc.), we target a discriminative feature space that is generalizable to both coarse- and fine-grained features. 

Today, the convolutional neural network (CNN) is arguably the most popular backbone in terms of designing ReID classifiers. However, the latest advances in deep learning have seen the great success of Transformers \cite{liu2021swin, zhou2021deepvit} in natural language processing and computer vision. This class of convolution-free models can be a better fit for the problem of object ReID using self-attention. Most existing research on object (Persons, Vehicles) ReID has been conducted for normal surveillance. Limited attention has been paid to object ReID in aerial surveillance.  Most approaches (e.g., part-based convolutional baseline \cite{sun2018beyond}, generative adaptive alignment network \cite{wang2019rgb}, multiple granularity network \cite{wang2018learning}) follow CNN-based architectures. Transformer \cite{dosovitskiy2020image} is a new trend in the solution of object ReID, establishing it as a strong baseline that beats current state-of-the-art methods.

Motivated by the recent success of the transformer in normal surveillance object ReID \cite{he2021transreid}, we adopt an all-attention transformer-based approach for UAV-based object Re-ID. To address the challenges arising from aerial images, we propose an uncertainty-aware approach that exploits hierarchical feature maps and global channel attention gate for object Re-identification in the UAV domain.  Our proposed approach uses a modified version of Pyramid Vision Transformer (PVT) \cite{wang2021pyramid} as the backbone, which is a convolution-free architecture. Then we apply spatial attention \cite{woo2018cbam} to these multi-resolution feature maps to put more focus on important features, filtering out irrelevant ones. To incorporate camera information, we add additional head to the PVT model and optimize the network using both camera ID, object ID, and center loss in a joint fashion. Moreover, we normalize the style variance present in different cameras, incorporating Batch Instance Normalization (BIN) \cite{nam2018batch} in our model. Finally, with the help of the channel attention aggregation gate \cite{zhou2019omni}, the model selectively learns the feature maps with higher weights. Inspired by recent work on modeling feature uncertainty for personal Re-ID \cite{yu2019robust}, we propose to model aerial uncertainty by predicting the variance of data as the model output. 

PVT-based object recognition with uncertainty estimation is particularly suitable for UAV-based Re-ID of people and vehicles in aerial surveillance and long-range biometrics \cite{ye2021deep}.
Our \textbf{technical contributions} can be summarized as follows: \textbf{(I)} We seek to explore a modified Pyramid Vision Transformer (PVT) tailored for object re-identification in UAV-based scenarios. The proposed model utilizes multiscale features to re-identify objects for aerial surveillance. \textbf{(II)} Spatial attention module helps focus on the relevant information of the feature maps by filtering out noise. The channel attention gate follows an adaptive fusion scheme, which dynamically selects the appropriate feature maps to exploit channel-wise dependencies. \textbf{(III)} We train our model considering uncertainty in terms of object identity and camera identity for multitask learning. The model is regularized using Batch Instance Normalization (BIN) to mitigate style variations in multiple cameras. \textbf{(IV)} Our proposed framework achieves state-of-the-art performance on two aerial surveillance datasets, PRAI-1581 \cite{zhang2020person} and VRAI \cite{wang2019vehicle}, respectively. 

\section{Methodology}
\label{sec:format}
\begin{figure*}[t]
   \begin{center}
   \includegraphics[width=\linewidth]{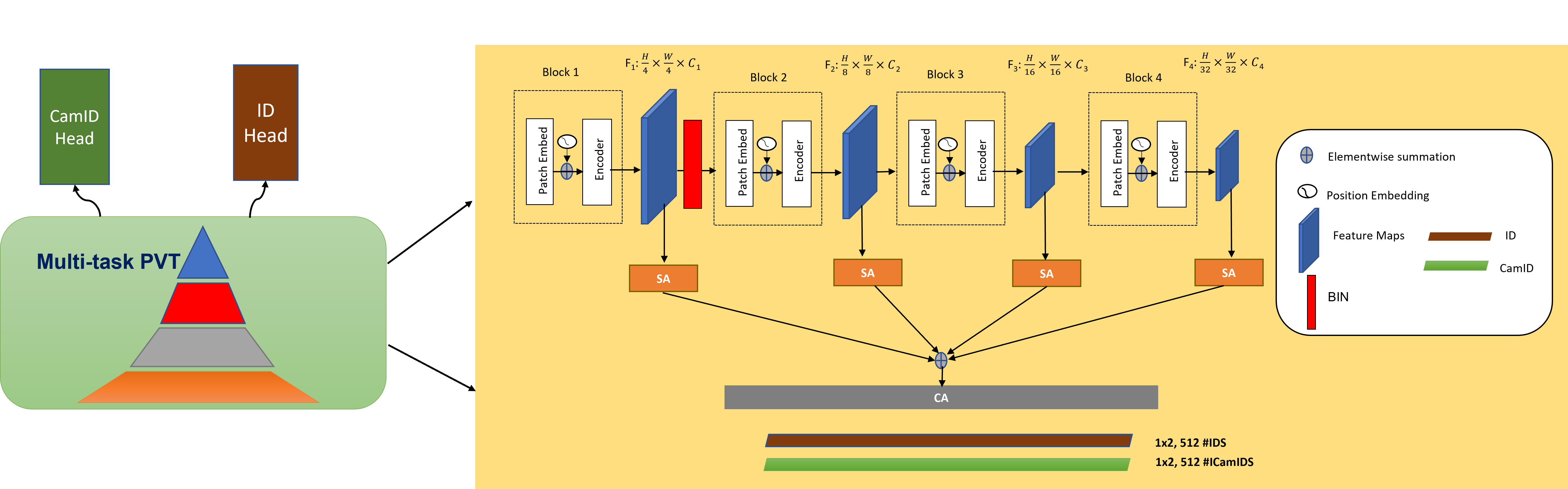}
   \end{center}
   \caption{The overall architecture of our proposed model. The model uses multi-task PVT as the backbone. We apply Spatial Attention (SA) on feature maps of individual scales which helps to focus the network on the most informative features. The channel attention (CA) is a mini-network with shared wights. The Identification (ID) head is responsible for classifying the object instances whereas CamID head gives prediction on camera ID labels considering uncertainty of prediction in both cases. Batch Instance Normalization (BIN) layer helps the network to achieve style generalization among different cameras.}
   \label{fig:model}
\vspace{-0.2in}
\end{figure*}

In this section, we present our multiscale approach for object Re-Identification. An overview of our proposed method is outlined in Figure \ref{fig:model}. We propose a multi-task Pyramid Vision Transformer (PVT) \cite{wang2021pyramid}, a convolution-free backbone designed to learn multiscale feature maps. We use two heads for object and camera ID recognition, respectively. Batch Instance Normalization (BIN) is incorporated in our model to achieve camera style invariance. To make feature maps spatially aware of the location of important objects, we apply spatial attention \cite{woo2018cbam} to feature maps of different resolutions. Finally, we combine multiscale feature maps in an adaptive way using a Channel Attention Gate \cite{woo2018cbam, zhou2019omni}. To make the identifier robust to occlusion, we estimate the aleatory uncertainty\cite{wang2019aleatoric} present in the data while computing the loss for the model.  
\subsection{Multi-task Pyramid Vision Transformer}
\label{sec:pagestyle}

Pyramid Vision Transformer (PVT) \cite{wang2021pyramid} generates hierarchical feature maps of multiple resolutions. This architecture has four blocks. Each block is responsible for generating a feature map of certain resolution. Each stage consists of a patch embedding layer and a transformer encoder layer. Each transformer encoder layer is composed of a modified attention layer named spatial reduction attention (SRA) and a feedforward layer. SRA is designed to reduce memory cost so that high-resolution feature maps can be processed.
First, the input image with a size of $H\times W\times3$ is separated into $\frac{HW}{4^2}$ patches and the patch size is $4\times4$. Then, each patch is flattened through a linear projection and passed through a number of transformer encoders. After that, we get the output feature map of size $\frac{H}{4}\times \frac{W}{4} \times C_1$. Similarly, we obtain the output feature maps of sizes $\frac{H}{8}\times \frac{W}{8} \times C_2$, $\frac{H}{16}\times \frac{W}{16} \times C_3$, and $\frac{H}{32}\times \frac{W}{32} \times C_4$, respectively.

The original PVT uses non-overlapping patches. Inspired by the recent work of TransReID \cite{he2021transreid}, we have generated overlapping patches using a sliding window or shifting operator.
The head in PVT is used to identify objects. Differently from the original PVT, we use an additional head for camera ID recognition. By doing this, we try to take advantage of the camera ID labels to fuse camera information into the original model. As the generated feature maps are produced by multiple transformer encoders, the generated feature space is complex, and it can be confusing for the model to extract meaningful features for the ReID task. To obtain a more refined feature map so that the model can learn more informative spatial features, we apply spatial attention to the feature maps of different resolutions. In spatial attention, both MaxPooling and AveragePooling are performed on the channel dimension, and the pooled feature space is concatenated to generate the 2D spatial attention map. 

Additionally, we have added Batch Instance Normalization (BIN) \cite{nam2018batch} to reduce style variations on multiple cameras. The purpose of using BIN is to normalize the style-preserving discriminative features of the ReID objectReID. To fully exploit the functionality of multiscale features, we combine feature maps of multiple resolutions using a global channel attention gate. To tune the channel weights of multiscale feature maps in a dynamic fashion, we use a shared channel attention gate to learn the inter-channel relationship within a feature map regarding the importance of feature maps. We follow the design procedure mentioned in \cite{zhou2019omni} to implement this gate as a mini-network composed of several global average pooling layers (GAP) and a multi-layer perceptron (MLP) with reduced hidden dimension and one ReLU-activated hidden layer followed by sigmoid activation. 

\subsection{Loss Function Formulation}
\label{sec:typestyle}
To train our model, we use a combination of three uncertainty-aware losses: identity loss,cameraID and center loss. Identity loss consists of softmax cross-entropy loss and triplet loss \cite{zhang2020rethinking,hermans2017defense}. Camera ID information is learned through centroid triplet loss \cite{wieczorek2021unreasonable} and center loss considers the distance from center information.\\
\textbf{Uncertainty-aware ID loss:} This loss is computed using a hybrid of softmax cross-entropy loss and triplet loss \cite{hu2018person} taking into account the uncertainty between classes. The uncertainty-aware softmax cross-entropy loss is calculated by \cite{zhang2020rethinking}:
\begin{equation}   
\label{e:id}
\mathcal{L}_{softmax}=\frac{1}{2N\sigma(x_i)^2}\sum_{i=1}^{N}log(p_{id}(h_i^{id},y_i))+ \frac{1}{2}{log\sigma(x_i)^2}
\end{equation}
where $N$ is the number of samples, $y_i$ is the ground truth label, and $\sigma(x_i)^2$ is the variance of the data. 
We model our uncertainty-aware soft-margin triplet loss using the following formula \cite{hermans2017defense}:
\begin{multline}   
\label{e:trip}
\mathcal{L}_{triplet}=\frac{1}{\sigma(x_a)^2}log(1+exp(f(x_a,x_n)-f(x_a,x_p))+\\ \frac{1}{2}{log\sigma(x_a)^2}
\end{multline}
where a triplet consists of $<x_a, x_p, x_n>$ in which $x_a$ is the anchor image of a person, $x_p$ is the positive anchor image belonging to the identity of the same person, and $x_n$ is the negative anchor image belonging to a different person. Note that triplet loss cannot measure the overall spatial distribution of features, while cross-entropy loss does not have enough discriminant power among features. Therefore, it is better to combine these two as follows:
\begin{equation}   
\label{e:id}
\mathcal{L}_{ua\_id}={L}_{softmax}+{L}_{triplet}
\end{equation}

\noindent\textbf{Uncertainty aware camera ID Loss:} 
To tackle intraclass variations arising from view angle, camera style, distance, etc., we apply a soft-margin version of centroid triplet loss since the class centroid can be considered as the mean representation for the retrieval task. Inspired by the unreasonable effectiveness of centroids in image retrieval \cite{wieczorek2021unreasonable}, we propose to calculate uncertainty-aware camera ID loss based on centroids as follows:
\begin{multline}
\mathcal{L}_{ua\_camid}=\frac{1}{\sigma(x_a)^2}log(1+exp(f(x_a,c_n)-f(x_a,c_p))+\\
\frac{1}{2}{log\sigma(x_a)^2},
\end{multline}
where $c_p$ and $c_n$ are the corresponding centroids of the class for the positive and negative classes. \\

\noindent\textbf{Uncertainty-aware Center Loss:} We also analyze uncertainty-aware center loss \cite{luo2019bag} using the following formula:
 \begin{equation}   
\label{e:tl}
\mathcal{L}_{ua\_center}=\frac{1}{2\sigma}\sum_{i=1}^{B} {\lvert\lvert {f}_{t_i}-{c_{y_i}}\rvert\rvert}^2_2,
\end{equation}
where $y_i$ is the label of the $i^th$ image in a mini-batch and B is the batch size. $c_{y_i}$ is the center of deep features in the $y_i$th class.

The overall loss can be formulated as follows.
 \begin{equation}   
\label{e:tl}
\mathcal{L}_{total}=\alpha_1\mathcal{L}_{ua\_id}+\alpha_2\mathcal{L}_{ua\_camid}+ \alpha_3 \mathcal{L}_{ua\_center}
\end{equation}
Here, $\alpha_1$, $\alpha_2$, and $\alpha_3$ are the regularization parameters for the corresponding losses.

\vspace{-0.2in}
\section{Experiments}
\label{sec:majhead}
\subsection{Datasets}
\vspace{-0.1in}
We have conducted our experiments on two aerial surveillance datasets named Person ReID for Aerial Imagery (PRAI-1581) \cite{zhang2020person} dataset and Vehicle Re-identification for Aerial Image (VRAI) \cite{wang2019vehicle} dataset. 
\textbf{PRAI} is a newly released aerial surveillance dataset which contains 39,461 person images of 1581 classes captured by two UAV drones with a flight altitude ranging from 20 to 60 meters above the ground. \textbf{VRAI dataset} consists around 137,613 images of 13,022 vehicles taken by two UAV drones. This is the largest UAV based vehicle dataset to date. 
\subsection{Implementation Details}
\vspace{-0.1in}
\label{ssec:subhead}
For the PRAI dataset, the training set includes
19,523 images from 782 classes. For the test set, the number of query and gallery images are 4680 and 15258, respectively. For the VRAI dataset, the training set contains 66,113 images with 6,302 classes. For the test set, the query set contains 15,747 images and the gallery set contains 55,753 images, respectively. 
In our experiment, we investigated PVT, a multiscale transformer, as the backbone network. The backbone network is pre-trained on ImageNet 2012 dataset. We train our model using 4 Titan 1080GTX GPUs. Before training, the images are resized to $224\times224$. The batch size is set to 128. ADAM optimizer is used with a momentum of 0.9 and a weight decay of $1e^{-4}$. The learning rate is initialized
as 0.000015 with a cosine rate decay. 
For performance evaluation, we use two metrics: Cumulative Matching Characteristic (CMC) and mean Average Precision (mAP).
\begin{figure}%
    \centering
    \subfloat[\centering Person]{{\includegraphics[width=24mm,height=18mm]{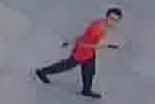} }}%
    \subfloat[\centering PVT-Large]{{\includegraphics[width=24mm,height=18mm]{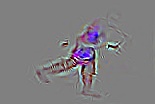} }}%
    \subfloat[\centering Ours]{{\includegraphics[width=24mm,height=18mm]{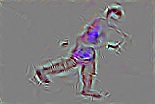} }}%
    \caption{Visualization of feature maps using guided back-propagation. Baseline PVT-Large \cite{wang2021pyramid} fails to retrieve fine features, while ours captures more discriminative features.}%
    \label{fig:example}%
    \vspace{-0.2in}
\end{figure} 
\subsubsection{Comparison with state-of-the-art}
\label{sssec:subsubhead}
\begin{figure}[htb]
\vspace{-0.2in}
\begin{minipage}[b]{.48\linewidth}
  \centering
  \centerline{\includegraphics[width=4.5cm,height=3.4cm]{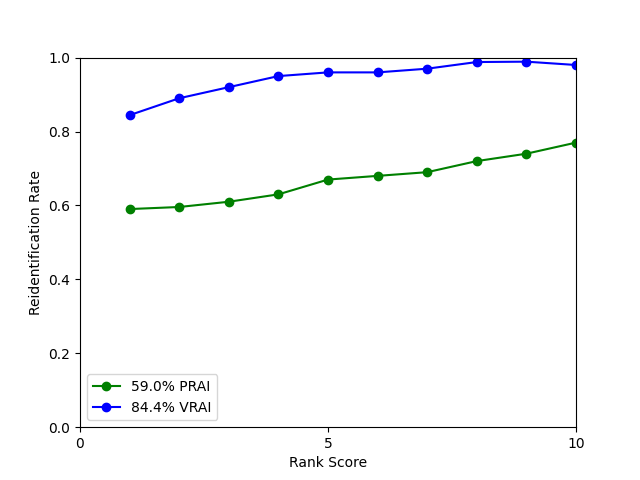}}
  \centerline{(a) CMC curve}\medskip
\end{minipage}
\hfill
\begin{minipage}[b]{0.48\linewidth}
  \centering
  \centerline{\includegraphics[width=4.5cm,height=3.2cm]{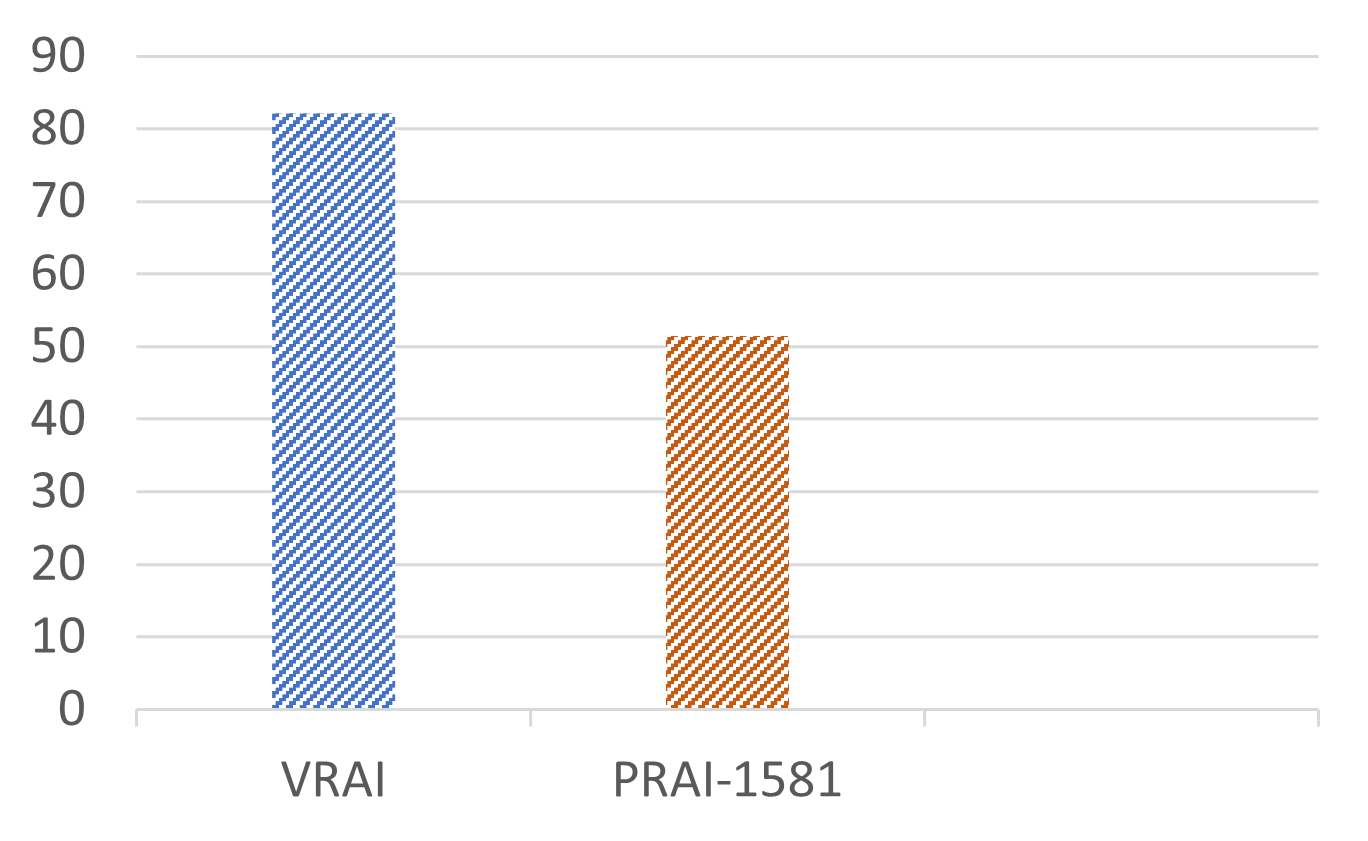}}
  \centerline{(b) mAP}\medskip
\end{minipage}
\caption{Performance of our model for PRAI-1581 and VRAI.}
 \vspace{-0.2in}
\label{fig:res}
\end{figure}

We compare our proposed approach with the latest methods, and the results are reported in Table \ref{t:bigtable}. Our proposed approach outperforms the previous state-of-the-art in both the PRAI-1581 and VRAI dataset, respectively. For the VRAI dataset, our approach achieves 84.47\% Rank-1 accuracy and an mAP of 82.86\%. For the PRAI dataset, the accuracy of Rank-1 is 59.18\% and the mAP is 51.45\%. We report results based on single-query settings for both datasets. It is worth mentioning that the gain over previous SOTA methods is consistent across object domains (person vs. vehicle) and performance metrics (Rank-1 vs. mAP).
The performances of our model for mAP and different rank scores are presented in Figure \ref{fig:res} for the PRAI-1581 and VRAI data set. It can be observed that PRAI is a lot more challenging than VRAI because of people's relatively smaller size, large pose variations, and deformable motion. 

\begin{table}
\caption{Performance comparison with state-of-the-art methods for PRAI-1581 and VRAI.}
\vspace{-0.2in}
\begin{center}
\begin{tabular}{l|c|c}
\label{t:bigtable}
Method (Person ReID)& Rank-1 & mAP \\
\hline\hline
PCB \cite{sun2018beyond,zhang2020person} & 47.47 &  37.15\\
\hline
SVDNet \cite{sun2017svdnet,zhang2020person}&  46.10& 36.70 \\
\hline
MGN \cite{wang2018learning,zhang2020person} &  49.64& 40.86 \\
\hline
OSNet \cite{zhou2019omni,zhang2020person}&  54.40& 42.10 \\
\hline
TransReID \cite{he2021transreid}&  56.30& 49.81 \\
\hline
\textbf{Ours} & \textbf{59.18} & \textbf{51.45}\\
\hline
Method (Vehicle ReID) & Rank-1 & mAP \\
\hline\hline
MGN \cite{wang2018learning,wang2019vehicle} & 67.84 & 69.49 \\
\hline
RAM (ResNet-50) \cite{liu2018ram,wang2019vehicle} & 68.58 & 69.37\\
\hline
RAM (VGG-16) \cite{liu2018ram,wang2019vehicle} & 72.05 & 57.33\\
\hline
Multi-task+DP \cite{wang2019vehicle} & 80.30 & 78.63 \\
\hline
TransReID \cite{he2021transreid}& 82.68 & 81.48 \\
\hline
\textbf{Ours} & \textbf{84.47} & \textbf{82.86}\\
\hline
\end{tabular}
\end{center}
\vspace{-0.2in}
\end{table}
\vspace{-0.2in}

\section{Conclusion}
\label{sec:print}

We have presented an uncertainty-aware multiscale transformer-based approach for the UAV-based object Re-ID. Our approach captures the information of instances with different levels of detail by multitasking PVT-based backbone architecture. The proposed model tries to solve object Re-ID as multitask learning problem using a unified framework trained with object ID, camera ID, and center loss. We quantitatively and qualitatively evaluated our proposed method on two UAV-based aerial surveillance datasets. The experimental results demonstrate the superiority of the proposed model over the previous state-of-the-art. 



\bibliographystyle{IEEEbib}
\bibliography{strings}

\end{document}